\documentclass[final]{cvpr}

\usepackage{times}
\usepackage{epsfig}
\usepackage{graphicx}
\usepackage{subfigure}
\usepackage{amsmath}
\usepackage{amssymb}

\usepackage[utf8]{inputenc} 
\usepackage[T1]{fontenc}    
\usepackage{url}            
\usepackage{booktabs}       
\usepackage{amsfonts}       
\usepackage{nicefrac}       
\usepackage{microtype}      
\usepackage{lipsum}
\usepackage{amsmath}
\usepackage{array}
\usepackage{float}
\usepackage{bm}
\usepackage{multirow}
\usepackage{booktabs}
\usepackage{subfigure}
\usepackage{algorithm}  
\usepackage{algorithmic}
\usepackage{graphicx}
\usepackage{xcolor}
\usepackage{float}
\usepackage{mathtools}

\def\E{{\rm E}}

\def\bz{\pmb{Z}}
\def\bx{\pmb{X}}
\def\bp{\pmb{P}}
\def\by{\pmb{Y}}

\DeclarePairedDelimiterX{\infdivx}[2]{(}{)}{%
  #1\;\delimsize\|\;#2%
}
\newcommand{\kl}{D_{KL}\infdivx}

\usepackage[pagebackref=true,breaklinks=true,colorlinks,bookmarks=false]{hyperref}

\def\cvprPaperID{5981} 
\def\confYear{CVPR 2021}

\begin{document}

\title{Trajectory Prediction with Latent Belief Energy-Based Model}

\author{Bo Pang, Tianyang Zhao, Xu Xie, and Ying Nian Wu\\
Department of Statistics, University of California, Los Angeles (UCLA)\\
{\tt\small \{bopang, tyzhao, xiexu\}@ucla.edu, ywu@stat.ucla.edu} \\
}

\maketitle

\begin{abstract}
Human trajectory prediction is critical for autonomous platforms like self-driving cars or social robots. We present a latent belief energy-based model (LB-EBM) for diverse human trajectory forecast. LB-EBM is a probabilistic model with cost function defined in the latent space to account for the movement history and social context. The low-dimensionality of the latent space and the high expressivity of the EBM make it easy for the model to capture the multimodality of pedestrian trajectory distributions.  LB-EBM is learned from expert demonstrations (i.e., human trajectories) projected into the latent space. Sampling from or optimizing the learned LB-EBM yields a belief vector which is used to make a path plan, which then in turn helps to predict a long-range trajectory. The effectiveness of LB-EBM and the two-step approach are supported by strong empirical results. Our model is able to make accurate, multi-modal, and social compliant trajectory predictions and improves over prior state-of-the-arts performance on the Stanford Drone trajectory prediction benchmark by $10.9\%$ and on the ETH-UCY benchmark by $27.6\%$.
\end{abstract}

\section{Introduction}

Forecasting the future trajectories of pedestrians is critical for autonomous moving platforms like self-driving cars or social robots with which humans are interacting. It has recently attracted interest from many researchers \cite{gupta2018social, zhao2019multi, lee2017desire, sadeghian2019sophie, bhattacharyya2019conditional, deo2020trajectory, liang2020simaug, mangalam2020not}. See \cite{rudenko2020human} for an overview. Trajectory forecast is a challenging problem since human future trajectories depend on a multitude of factors such as past movement history, goals, behavior of surrounding pedestrians. Also, future paths are inherently multimodal. Given the past trajectories, there are multiple possible future paths. We propose a latent belief energy-based model (LB-EBM) which captures pedestrian behavior patterns and subtle social interaction norms in the latent space and make multimodal trajectory predictions. LB-EBM is learned from expert demonstrations (i.e., human trajectories) following the principle of inverse reinforcement learning (IRL) \cite{ng2000algorithms, finn2016connection, finn2016guided, haarnoja2017reinforcement}.

Traditional IRL approaches \cite{ng2000algorithms} first learn a cost function from expert demonstrations in an outer loop and then use reinforcement learning to extract the policy from the learned cost function in an inner loop. These approaches are often highly computationally expensive. To avoid such an issue, GAIL (Generative Adversarial Imitation Learning) \cite{ho2016generative, deo2020trajectory} optimizes a policy network directly. GAIL can generate multimodal action predictions given an expressive policy generator. The multimodality is however modeled implicitly and completely relies on the policy generator. Our approach strikes a middle ground between traditional IRL and GAIL. We learn an energy-based model (EBM) as the cost function in a low dimensional latent space and map the EBM distribution to actions with a policy generator. Similar to traditional IRL, we learn a cost function but our cost function is defined in a low dimensional space so that our cost function is easier to model and learn. Resembling GAIL, we also learn a policy generator which allows for directly mapping a latent vector to the action trajectory, while we explicitly learn a multimodal cost function instead of learning it implicitly and completely relying on the policy generator.

An EBM \cite{xie2016theory, nijkamp2019learning, pang2020learning} in the form of Boltzmann or Gibbs distribution maps a latent vector to its probability. It has no restrictions in its form and can be instantiated by any function approximators such as neural networks. Thus, this model is highly expressive and learning from human trajectories allows it to capture the multimodality of the trajectory distribution. Our proposed LB-EBM is defined in a latent space. An encoder is jointly learned to project human trajectories into the latent space and hence provides expert demonstrations to the latent cost function.

Furthermore, this cost function accounts for trajectory history and motion behavior of surrounding pedestrians.  Thus sampling from or optimizing the cost function yields a latent belief, regarding future trajectory, which considers the centric agent's behavior pattern and social context surrounding this agent. A future trajectory is then forecasted in two steps. We first use the social-aware latent belief vector to make a rough plan for future path. It is intuitive that human do not plan every single future step in advance but we often have a rough idea about how to navigate through our future path, which is based on one's belief after observing other agents' motion. The belief is inherently related to the agent's behavior pattern. This forms the intuitive motivation of our modeling approach. Conditioned on the plan, the trajectory is then predicted with the assistance of individual motion history and social cues. Several recent works take two steps to make trajectory forecast. They either first estimate the final goal \cite{mangalam2020not} or make a plan on a coarse grid map \cite{liang2020garden}. We take a similar approach. The plan in our approach is defined to be positions of some well-separated steps in the future trajectory, which can be easily extracted from the data.

The proposed LB-EBM and other modules are learned end-to-end. We test our model on the Stanford Drone (SDD) trajectory prediction benchmark and the ETH-UCY benchmark and improves the prior state-of-the-art performance by  $10.9\%$ on SDD and $27.6\%$ on ETH-UCY.\\ 

Our work has the following contributions. 
\begin{itemize}
	\item We propose a latent belief energy-based model (LB-EBM), following the principle of IRL, which naturally captures the multimodal human trajectory distribution.
	\item Our approach predicts multimodal and social compliant future trajectories.
	\item Our model achieves the state-of-the-art on widely-used human trajectory forecasting benchmarks.
\end{itemize}

\section{Related Work} 

Agents’ motions depend on their histories, goals, social interactions with other agents, constraints from the scene context, and are inherently stochastic and multimodal. Conventional methods of human trajectory forecasting model contextual constraints by hand-crafted features or cost functions \cite{vehicle, social_force, who_are_you}. With the recent success of deep networks, RNN-based approaches have become prevalent. These works propose to model interactions among multiple agents by applying aggregation functions on their RNN hidden states \cite{alahi2016social, gupta2018social, head}, running convolutional layers on agents' spatial feature maps \cite{conv_social_pooling, dwivedi2020ssp, zhao2019multi, Wu_2020_CVPR}, or leveraging attention mechanisms or relational reasoning on constructed graphs of agents \cite{li2020Evolvegraph, sadeghian2019sophie, carnet, zhang2019sr, social_attention}. Some recent studies are, however, rethinking the use of RNN and social information in modeling temporal dependencies and borrowing the idea of transformers into the area \cite{giuliari2020transformer}. We apply these social interaction modeling approaches with a few modifications in our work.

\textbf{Modeling Goals}. Recent progress has suggested that directly modeling goals could significantly decrease the error for trajectory forecasting. \cite{Rhinehart_2019_ICCV} introduces a prediction method conditioning on agent goals. \cite{mangalam2020not} proposes to first predict the goal based on agents' individual histories and then to forecast future trajectories conditioning on the predicted goal. \cite{liang2020garden} introduces a two-step planning scheme, first in a coarse grid then in a finer one, which can be viewed as directly modeling goals and sub-goals. We follow the general scheme of two-step prediction. The plan in our approach is defined to be positions of some well-separated steps in the future trajectory, which can be easily extracted from the data.

\textbf{Multimodality}. Most recent prediction works have emphasized more on modeling the multimodality nature of human motions. \cite{interaction_maneuver, maneuver} directly predict multiple possible maneuvers and generate corresponding future trajectories given each maneuver. \cite{lee2017desire, ivanovic2019trajectron} use Variational Auto-Encoders \cite{vae} and \cite{gupta2018social, lee2017desire, sadeghian2019sophie, zhao2019multi} use Generative Adversarial Networks \cite{gan, cgan} to learn distributions. Many works \cite{liang2020garden, Phan-Minh_2020_CVPR, r2p2, tang2019mfp} also focus on developing new datasets, proposing different formulations, utilizing latent variable inference, and exploring new loss functions to account for multimodality. Our work adopts the likelihood-based learning framework with variational inference. We propose a novel way to model the multimodality of human trajectories, by projecting them into a latent space with variational inference and leveraging the strength of latent space energy-based model.

\textbf{Value Function}. Human behaviors are observed as actions, e.g. trajectories, but the actions are actually guided by hidden value functions, revealing human preference and cost over different actions. Some previous works explicitly or implicitly model these types of cost functions as intermediate steps for sampling possible futures. These works generally follow the reinforcement learning formulation of value functions $Q$. \cite{Ngai2011AMR} directly uses Q-Learning to learn value functions. \cite{xu2019ioc, gail_gru} formulate trajectory planning and prediction problems as inverse optimal control and GAIL (generative adversarial imitation learning) problems. \cite{fictitious} models social interaction by game theory and attempt to find the hidden human value by fictitious play. P2TIRL \cite{deo2020trajectory} is learned by a maximum entropy inverse reinforcement learning (IRL). Our work also follows the basic principle of inverse reinforcement learning to learn human cost functions explicitly in a latent space.

\textbf{Energy-Based Models}. The energy function in the EBM \cite{zhu1998frame, xie2016theory, nijkamp2019learning, du2019implicit, han2020joint} can be viewed as an objective function, a cost function, or a critic  \cite{sutton2018reinforcement}. It captures regularities, rules or constrains. It is easy to specify, although optimizing or sampling the energy function requires iterative computation such as MCMC. Recently \cite{pang2020learning, pang2020molecule, pang2020semi} proposed to learn EBM in a low dimensional latent space, which makes optimizing or sampling the energy function much more efficient and convenient. The current work follows this approach.

\section{Model and Learning}

\subsection{Problem Definition}
Let $x_i^t \in \mathbb{R}^2$ denote the position of a person $i$ at time $t$ in a scene where there are $n$ people in total. The history trajectory of the person $i$ is $\pmb{x}_i = \{x_i^t, t=1,..., t_{past}\}$ and $\pmb{X} = \{\pmb{x}_i, i = 1,...,n\}$ collects past trajectories of all people in a scene. Similarly, the future trajectory of this person  at time $t$ is denoted as $y_i^t$. $\pmb{y}_i = \{y_i^t, t=t_{past}+1,..., t_{pred}\}$ and $\pmb{Y} = \{\pmb{y}_i, i = 1,...,n\}$ indicate the future trajectory of the person $i$ and all future trajectories, respectively. The goal is to jointly predict the future trajectories of all the agents in the scene or to learn the probabilistic distribution, $p(\pmb{Y} | \pmb{X})$.

\begin{figure*}[t!]
    \centering
    \includegraphics[width=0.7\linewidth]{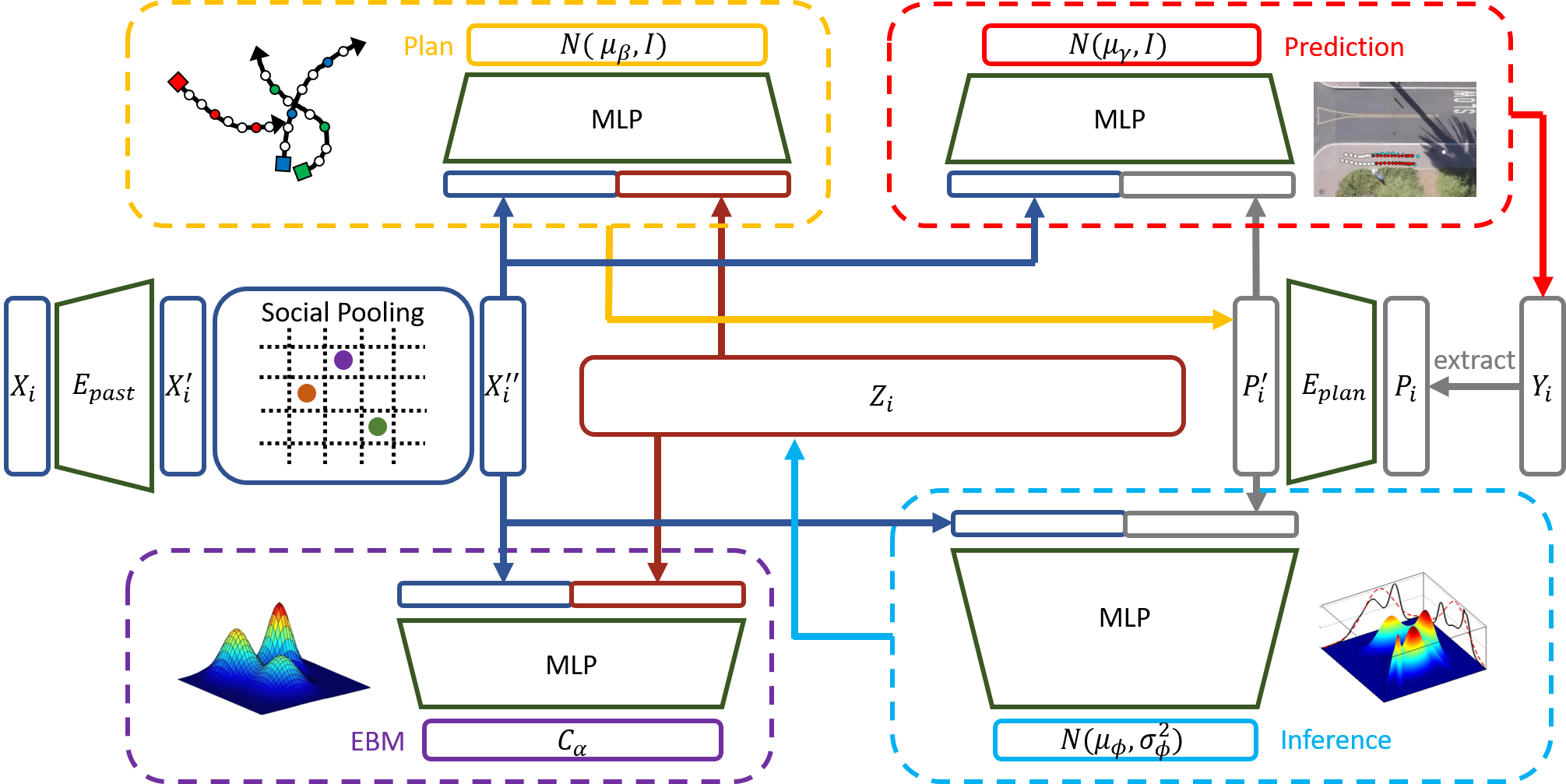}
    \caption{An overview of our model on an individual agent $i$. The past trajectory $x_i$ (left side in the figure) is encoded by $E_{past}$ to get the individual encoding $x'_i$. The social pooling module $P_{social}$ is then applied to get the agent's history encoding $x''_i$ accounting for social context. In training, the ground-truth plan $p_i$ (right side in the figure) is extracted from the future trajectory $y_i$ (e.g., extract the steps 3, 6, 9, 12 from a 12-time-step future as the plan) and then encoded by $E_{plan}$ to get $p'_i$. The expert plan is then projected into the latent space, conditional on the trajectory history and social context, $x''_i$, through the inference module (light blue). It takes $x''_i$ and $p'_i$ as input, parameterized by $\phi$, and is only used in training to output the mean $\mu_{\phi}$ and co-variance matrix $\sigma_{\phi}^2$ for the posterior distribution, $q_{\phi}$, of the latent vector $z_i$. Purple part denotes the latent belief energy-based model (LB-EBM) module, $C_\alpha$, defined on the latent belief vector $z_i$ conditional on $x''_i$. The LB-EBM learns from the posterior distribution of the projected ground-truth plan $q_{\phi}$. A sample from the posterior (in training) or a sample from LB-EBM (in testing) enters the plan module (yellow) together with $x''_i$. The plan module is parametrized by $\beta$, which is a regular regression model where the mean $\mu_\beta$ is estimated and used as the module prediction. The generated plan together with $x''_i$ enters the prediction module (red), parameterized by $\gamma$. It is also a regular regression model where the mean $\mu_\gamma$ is estimated and used as the module prediction, which is also the trajectory forecast of the whole network.}
    \label{fig:ebm}
\end{figure*}

Directly modeling $p(\pmb{Y} | \pmb{X})$ is essentially supervised learning or behavior cloning which often fails to capture the multimodality. Instead, we introduce two auxiliary variables. The first is $\pmb{z}_i$ which represents the latent belief of the agent $i$ after observing the trajectory history of his or her own and surrounding agents, $\pmb{X}$. Let $\pmb{Z}=\{\pmb{z}_i, i=1,...,n\}$. $\pmb{z}_i$ is a latent variable since we cannot observe one's latent belief. The other auxiliary variable is $\pmb{p}_i$ which denotes the plan of the agent $i$ considering the latent belief $\pmb{z}_i$  and trajectory history $\pmb{X}$. Similarly, let $\pmb{P}=\{\pmb{p}_i, i=1,...,n\}$. $\pmb{p}_i$ can be either latent or observed. We choose to use a few well-separated steps of future trajectory, $\pmb{y}_i$, to represent one's plan, making it an observable. Thus, we can extract plan from the data to provide supervision signal, making the learning easier. With the aforementioned setup, we model the following joint distribution,
\begin{equation}
p(\pmb{Z}, \pmb{P}, \pmb{Y} | \pmb{X}) = \underbrace{p(\pmb{Z} | \pmb{X})}_\text{LB-EBM} \overbrace{p(\pmb{P} |\pmb{Z}, \pmb{X})}^\text{Plan} \underbrace{p(\pmb{Y} |\pmb{P},\pmb{X})}_\text{Prediction}. 
\end{equation}

After learning the model, we can follow the above chain to make trajectory prediction. A well-learned LB-EBM or cost function captures expert's belief distribution given trajectory history and motion behavior of surrounding agents. Sampling from or optimizing this cost function gives a good belief representation taking account into individual behavior pattern and social context. This cost function is inherently multimodal since it learns from the multimodal human trajectories. We can then make a plan with $p(\pmb{P} |\pmb{Z}, \pmb{X})$ (the plan module) by directly generating a trajectory plan. Lastly, $p(\pmb{Y} |\pmb{P},\pmb{X})$ (the prediction module) makes a trajectory prediction given the plan and past history. In the following section, we detail each part of the decomposed distribution and introduce related encoding functions.

\subsection{LB-EBM}

In our approach, the key step is to learn a cost function defined in a latent belief space. For a latent belief vector $\pmb{z}_i$, the cost function is defined to be 
\begin{equation}
C_{\alpha}(\pmb{z}_i, P_{social}(\pmb{X}))    
\end{equation}
where $\alpha$ denotes the parameters of the cost function. Two relevant encoding modules are, $E_{past}$ which is used to encode the trajectory history $\pmb{x}_i$ of each agent and $P_{social}$ which is a pooling module that aggregates $\{E_{past}(\pmb{x}_i), i=1,...,n\}$ to provide the latent belief space with individual behavior history and social context. $C_{\alpha}(\cdot)$ takes $[\pmb{z}_i; P_{social}(\pmb{X})]$ as the input where $[\ \cdot \ ; \cdot \ ]$ indicates concatenation.

Assuming we have a well-learned cost function, we can find a $\pmb{z}_i$ by minimizing the cost function with respect to it given $\pmb{X}$, generate a plan with the latent belief, and then make the trajectory plan. The cost function is learned from expert demonstrations projected into the latent space. A plan, $\pmb{p}_i$, extracted from an observed future human trajectory, $\pmb{y}_i$, can be projected to the latent space. Suppose $\pmb{y}_i$ consists of $12$ time steps and $\pmb{p}_i$ can take the positions at the $3$rd, $6$th, $9$th, and $12$th time steps as the plan. Denote the projected latent vector to be $\pmb{z}_i^+$. $\alpha$ is learned from $\{\pmb{z}_{ij}^+, i=1,...,n; j=1,...,N\}$ where $j$ indicates the $j$th scene with $N$ scenes in total. See section \ref{learning} for the learning details. The projection or inference is done by an inference network $E_{inference}$. The distribution of the inferred latent belief is $q_{\phi}(\pmb{z}_i |\pmb{p}_i, \pmb{X})$, which is assumed to be a multivariate Gaussian with a diagonal covariance matrix. In particular, the mean function $\mu_{\phi}(\pmb{p}_i, \pmb{X})$ and covariance matrix $\sigma^2_{\phi}(\pmb{p}_i; \pmb{X})$ both takes $[E_{plan} (\pmb{p}_i); P_{social}(\pmb{X})]$ as the input and share the neural network module except the last layer. Here $E_{plan}$ is simply an embedding function which encodes the plan $\pmb{p}_i$ into a feature space to be ready to concatenate with $P_{social}(\pmb{X})$. 

The LB-EBM assumes the following conditional probability density function
\begin{align}
&p_{\alpha}(\pmb{z}_i | P_{social}(\pmb{X})) \\
&= \frac{1}{Z_{\alpha}(P_{social}(\pmb{X}))} \exp{[-C_{\alpha}(\pmb{z}_i, P_{social}(\pmb{X}))]}p_0(\pmb{z}_i),
\end{align}
where $Z_{\alpha}(P_{social}(\pmb{X})) = \int \exp{[-C_{\alpha}(\pmb{z}_i, P_{social}(\pmb{X}))]} d \pmb{z}_i$ is the normalizing constant or partition function and $p_0(\pmb{z}_i)$ is a known reference distribution, assumed to be standard Gaussian in this paper. The cost function $C_{\alpha}$ serves as the energy function. The latent belief vectors of experts $\pmb{z}_{ij}^+$ are assumed to be random samples from $p_{\alpha}(\pmb{z}_i | P_{social}(\pmb{X}))$ and thus has low cost on $C_{\alpha}(\pmb{z}_i, P_{social}(\pmb{X}))$.

The joint distribution of the latent belief vectors of agents in a scene is then defined to be
\begin{align}
p(\pmb{Z} | \pmb{X}) = \prod_{i=1}^n p_{\alpha}(\pmb{z}_i | P_{social}(\pmb{X})),
\end{align}
where $\{\pmb{z}_i, i=1,...,n\}$ given the joint trajectory history $\pmb{X}$ are independent because an agent cannot observe the belief of other agents.

To sample from LB-EBM, we employ Langevin dynamics \cite{neal2011mcmc, zhu1998grade, nijkamp2020learning}. For the target distribution $p_{\alpha}(\pmb{z} | P_{social}(\pmb{X}))$, the dynamics iterates 
\begin{align} 
\pmb{z}_{k+1} = \pmb{z}_k + s \nabla_{\pmb{z}} \log p_{\alpha}(\pmb{z} | P_{social}(\pmb{X})) + \sqrt{2s} \epsilon_k, 
\label{eq:Langevin}
\end{align}
where $k$ indexes the time step of the Langevin dynamics, $s$ is a small step size, and $\epsilon_k \sim {\rm N}(0, I)$ is the Gaussian white noise. Note that the index $i$ for $\pmb{z}$ is removed for notational simplicity. $\nabla_{\pmb{z}} \log p_{\alpha}(\pmb{z} | P_{social}(\pmb{X}))$ can be efficiently computed by back-propagation. Given the low-dimenionality of the latent space, Langevin dynamics sampling mixes fast. In practice, we run the dynamics for a fixed number of times ($20$). The small number of steps and the small model size of the LB-EBM make it highly affordable in practice.

\subsection{Plan}
The distribution of the plan of the agent $i$ is $p_{\beta}(\pmb{p}_i |\pmb{z}_i, \pmb{X} )$, and it is assumed to be a Gaussian distribution with mean $\mu_{\beta}(\pmb{z}_i, \pmb{X})$ and an identity covariance matrix. In particular the mean function takes as input the concatenation $[\pmb{z}_i; P_{social}(\pmb{X})]$. The joint distribution of the plans of all agents in a scene is 
\begin{align}
p(\pmb{P} |\pmb{Z}, \pmb{X}) = \prod_{i=1}^n p_{\beta}(\pmb{p}_i |\pmb{z}_i, P_{social}(\pmb{X})),
\end{align}
where $\pmb{p}_i$ is assumed to be independent of $\{ \pmb{z}_j, j \neq i \}$ given $\pmb{z}_i$ and $P_{social}(\pmb{X})$ and $\{\pmb{p}_i, i=1,...,n\}$ are assumed to be independent conditional on $\{ \pmb{z}_i \}$ and $P_{social}(\pmb{X})$.  

\subsection{Prediction}
The prediction distribution is defined similarly as the plan distribution,
\begin{align}
p(\pmb{Y} |\pmb{P}, \pmb{X}) = \prod_{i=1}^n p_{\gamma}(\pmb{y}_i |\pmb{p}_i, P_{social}(\pmb{X})),
\end{align}
and $p_{\gamma}(\pmb{y}_i |\pmb{p}_i, P_{social}(\pmb{X}))$ assumes a Gaussian distribution with mean $\mu_{\gamma}(\pmb{p}_i, \pmb{X})$ and an identity covariance matrix. The input to the mean function is $[E_{plan}(\pmb{p}_i); P_{social}(\pmb{X})]$.

\subsection{Pooling}
The trajectory history $\pmb{X}$ of agents in a scene is pooled through self-attention \cite{vaswani2017attention}. It allows us to enforce a spatial-temporal structure on the social interactions among agents. This enforcement is simply achieved by designing a spatial-temporal binary mask with prior knowledge. We follow the mask design of \cite{mangalam2020not}. The pooling mask $M$ is defined to be,
\begin{equation}
    M[i,j] =
    \begin{cases}
      0 & \text{if}\ \displaystyle\min_{1 \leq s,t \leq t_{past}} \Vert x_i^t - x_j^s \Vert_2 > d \\
      1 & \text{otherwise}.
    \end{cases}
\end{equation}
Adjusting the hyperparameter $d$ allows for varying the social-temporal adjacency of social interactions.

\subsection{Joint learning}
\label{learning}
The log-likelihood of data in a single scene, $(\pmb{X}, \pmb{Y}, \pmb{P})$, is 
\begin{align}
\log p(\pmb{P}, \pmb{Y} | \pmb{X}) &= \log \int_{\pmb{Z}} p(\pmb{Z}, \pmb{P}, \pmb{Y} | \pmb{X})\label{eq:mle} 
\end{align}
which involves the latent variable $\pmb{Z}$ and directly optimizing it involves sampling from the intractable posterior $p(\pmb{Z}|\pmb{P}, \pmb{X})$. We however can optimize a variational lower bound of it in an end-to-end fashion to learn the entire network,
\begin{align}
L(\theta) &= \E_{q_{\phi}(\pmb{Z} |\pmb{P}, \pmb{X})} \log p_{\beta}(\pmb{P} |\pmb{Z}, \pmb{X}) \label{eq:recon}\\ 
&+ \E_{q_{data}(\pmb{Y} | \pmb{P}, \pmb{X})} \log p_{\gamma}(\pmb{Y} |\pmb{P}, \pmb{X})  \label{eq:pred}\\
& - \mathbb{KL}(q_{\phi}(\pmb{Z} |\pmb{P}, \pmb{X}) || p_0(\pmb{Z})) \label{eq:kl1}\\
& - \E_{q_{\phi}(\pmb{Z} |\pmb{P}, \pmb{X})}C_{\alpha}(\pmb{Z}, \pmb{X}) - \log Z_{\alpha}(\pmb{X}) \label{eq:kl2},
\end{align}
where $\theta$ collects the parameters of the whole network. Also note that $p_0(\pmb{Z})=\prod_i p_0(\pmb{z}_i)$ and $C_{\alpha}(\pmb{Z}, \pmb{X}) = \sum_i C_{\alpha}(\pmb{z}_i, \pmb{X})$. Please see the supplementary for the derivation details. The gradients of all terms are straightforward with backpropagation except $ \log Z_{\alpha}(\pmb{X})$. The gradient of it with respect to $\alpha$ is $\E_{p(\pmb{Z} | \pmb{X})}[\nabla_{\alpha} C_{\alpha}(\pmb{Z}, \pmb{X})]$. It involves sampling from LB-EBM. This is done with Langevin dynamics (Equation \ref{eq:Langevin}). As we discussed earlier, sampling from LB-EBM only requires a small number of steps and the necessary model size is fairly small due to the low dimensionality. Thus the sampling is highly affordable. Although the loss function $- L\{\theta\}$ is optimized end-to-end, let us take a close look at the optimization of the cost function given its core role in our model. Let $\mathcal{J}(\alpha)$ be the loss function of the LB-EBM, the gradient of it with respect to $\alpha$ is,
\begin{align}
&\nabla_{\alpha} \mathcal{J}(\alpha)\\
&= \E_{q_{\phi}(\pmb{Z} |\pmb{P}, \pmb{X})} [\nabla_{\alpha} C_{\alpha}(\pmb{Z}, \pmb{X})] - \E_{p(\pmb{Z} | \pmb{X})}[\nabla_{\alpha} C_{\alpha}(\pmb{Z}, \pmb{X})],
\end{align}
where $q_{\phi}(\pmb{Z} |\pmb{P}, \pmb{X})$ projects the expert plan $\pmb{P}$ to the latent belief space. $\alpha$ is updated based on the difference between the expert beliefs and those sampled from the current LB-EBM. Thus, the latent cost function is learned to capture expert beliefs given the trajectory history and surrounding context.

\section{Experiments}
We test our model on two widely used pedestrians trajectory benchmarks (see section \ref{datasets} for details) against a variety competitive baselines. These experiments highlight the effectiveness of our model with (1) improvements over the previous state-of-the-art models on the accuracy of trajectory prediction and (2) the prediction of multimodal and social compliant trajectories as demonstrate in qualitative analysis. 

\subsection{Implementation Details and Design Choices}
The trajectory generator or policy network is an autoregressive model in most prior works \cite{alahi2016social, gupta2018social, lee2017desire, sadeghian2019sophie}. Some recent works explored the use of a non-autoregressive model \cite{mangalam2020not, qi2020imitative}. We choose to use a non-autoregressive model (MLP) considering its efficiency and the avoidance of exposure bias inherent in autoregressive models. The potential issue of using an non-autoregressive model is that it might fail to capture the dependency among different time steps. However, this is a lesser issue since the proposed LB-EBM is expressive and multi-modal and might be able to model the dependency across multiple time steps. Furthermore, the trajectory prediction is based on a plan over the whole forecasting time horizon, making an auto-regressive model further unnecessary. 

The dimension of LB-EBM is $16$ and is implemented with 3-layer MLP with an hidden dimension of $200$. We always use 20 steps for Langevin sampling from LB-EBM in both training and inference. It is possible to amortize the sampling on the learned cost function by learning an auxiliary latent generator such as using noise contrastive estimation \cite{gutmann2010noise}. However, due to the low dimensionality of the latent space, 20 steps are highly affordable. We thus prefer keeping our model and learning method pure and simple.  

In both benchmarks, the model aims to predict the future 12 time steps. The plan is extracted by taking the positions at the $3$rd, $6$th, $9$th, and $12$th time steps.

All other modules in our model are also implemented with MLPs. The batch size is $512$ for the Stanford Drone dataset and is $70$ for all the ETH-UCY datasets. The model is trained end-to-end with an Adam optimizer with an learning rate of $0.0003$. The model is implemented in Pytorch \cite{NEURIPS2019_9015}. Our code is released at \url{https://github.com/bpucla/lbebm}.

\subsection{Datasets}
\label{datasets}

\textbf{Stanford Drone Dataset}. Stanford Drone Dataset \cite{robicquet2016learning} is a large-scale pedestrian crowd dataset in bird's eye view. It consists of 20 scenes captured using a drone in top down view around the university campus containing several moving agents such as humans bicyclists, skateboarders and vehicles. It consists of over $11,000$ unique pedestrians capturing over $185,000$ interactions between agents and over $40,000$ interactions between the agent and scene \cite{robicquet2016learning}. We use the standard train-test split which is widely used in prior works such as \cite{sadeghian2019sophie,gupta2018social,mangalam2020not}. 

\textbf{ETH-UCY}. It is a collection of relatively small benchmark pedestrian crowd datasets. It consists of five different scenes:  ETH and HOTEL (from ETH) and UNIV, ZARA1, and ZARA2 (from UCY). The positions of pedestrians are in world-coordinates and hence the results are reported in meters. We use the leave-one-out strategy for training and testing, that is, training on four scenes and testing on the fifth one, as done in previous works \cite{gupta2018social,li2019conditional, mangalam2020not}. We split the trajectories into segments of 8s and use 3.2s of trajectory history and a 4.8s prediction horizon, with each time step of 0.4s.

\subsection{Baseline Models}
We compare the proposed approach based on LB-EBM to a wide range of baseline models and state-of-the-art works. The compared work covers very different learning regimes for modeling human trajectory and accounting for multimodality and social interaction. We briefly describe below the representative baselines. 

\begin{itemize}
\item S-LSTM \cite{alahi2016social} is the simplest deterministic baseline based on social pooling on LSTM states.
\item S-GAN-P \cite{gupta2018social} is a stochastic GAN-based simple baseline extended from S-LSTM.
\item MATF \cite{zhao2019multi} is a GAN-based convolutional network built upon feature maps of agents and context.
\item Desire \cite{lee2017desire} is an VAE-based sophisticated stochastic model.
\item Sophie \cite{sadeghian2019sophie} is a complex attentive GAN modeling both social interactions and scene context.
\item CGNS \cite{li2019conditional} uses conditional latent space learning with variational divergence minimization. \item P2TIRL \cite{deo2020trajectory} is learned by maximum entropy inverse reinforcement learning policy. 
\item SimAug \cite{liang2020simaug} uses additional 3D multi-view simulation data adversarially. 
\item PECNet \cite{mangalam2020not} is a VAE based state-of-the-art model with goal conditioning predictions.
\end{itemize}

\subsection{Quantitative Results}
In this section, we compare and discuss our method's performance against the aforementioned baselines based on the Average Displacement Error (ADE) and Final Displacement Error (FDE) with respect to each time-step $t$ within the prediction horizon. 
\begin{equation}
  \begin{aligned}
{\rm ADE}_i & = \frac {1}{T_{\rm pred}} \sum_ {t = 1}^{T_{\rm pred}} d_{l_2} (\hat y_{i}^t , y_{i}^t) \quad \\
{\rm ADE} & = \frac{1}{n} \sum_i {\rm ADE}_i \\
{\rm FDE}_i & = d_{l_2}(\hat y_{i}^{T_{\rm pred}} , y_{i}^{T_{\rm pred}}) \quad \\
{\rm FDE} & = \frac{1}{n} \sum_i {\rm FDE}_i \\
 \end{aligned}
\end{equation}
where $d_{l_2}$ indicates the Euclidean distance. Following the evaluation protocol of the prior work \cite{gupta2018social, kosaraju2019social, mangalam2020not, zhao2019multi}, we use Best-of-K evaluation. In particular, the minimum ADE and FDE from $K$ randomly sampled trajectories are considered as the model evaluation metrics. And $K=20$ is used in our experiments. Recently, some researchers \cite{ivanovic2019trajectron, salzmann2020trajectron++, thiede2019analyzing} propose to use kernel density estimate-based negative log likelihood (KDE NLL) for evaluation. Since only few papers reported NLL results on our considered benchmarks and thus it might not be easy to have a fair comparison with most baselines, we choose to focus on the widely-adopted ADE and FDE. Please see the supplementary for the NLL evaluation of our model.

\begin{table}[!htbp]
\centering
\begin{tabular}{c||c|c}
 &  ADE & FDE \\\hline
\midrule
S-LSTM~\cite{alahi2016social}  & 31.19 & 56.97 \\\hline
S-GAN-P~\cite{gupta2018social}  & 27.23 & 41.44 \\\hline
MATF~\cite{zhao2019multi}  & 22.59 & 33.53 \\\hline
Desire~\cite{lee2017desire}  & 19.25 & 34.05 \\\hline
SoPhie~\cite{sadeghian2019sophie}  & 16.27 & 29.38 \\\hline
CF-VAE~\cite{bhattacharyya2019conditional}  & 12.60 & 22.30 \\\hline
P2TIRL~\cite{deo2020trajectory}  & 12.58 & 22.07 \\\hline
SimAug~\cite{liang2020simaug}  & 10.27 & 19.71 \\\hline
PECNet~\cite{mangalam2020not} &9.96 &15.88 \\\hline

\midrule
\textbf{Ours} &\textbf{8.87} & \textbf{15.61} \\\hline

\end{tabular}
\caption{ADE / FDE metrics on Stanford Drone for LB-EBM compared to baselines are shown. All models use 8 frames as history and predict the next 12 frames. The lower the better.}
\label{tab:sdd}
\end{table}

\begin{figure*}[t!]
    \centering
    \subfigure{\includegraphics[width=0.24\linewidth]{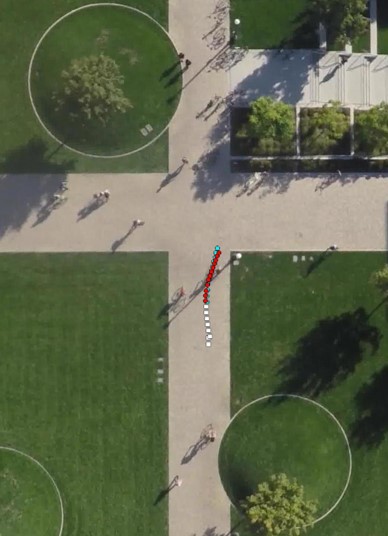}}
    \hfill
    \subfigure{\includegraphics[width=0.24\linewidth]{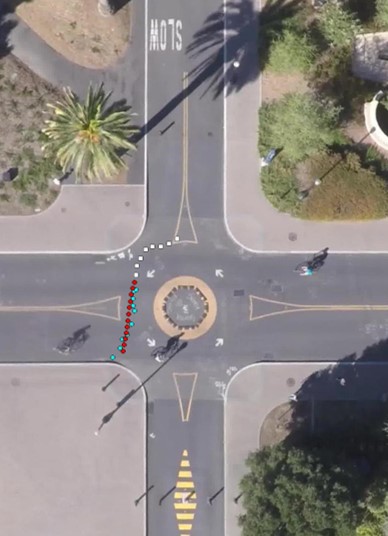}}
    \hfill
    \subfigure{\includegraphics[width=0.24\linewidth]{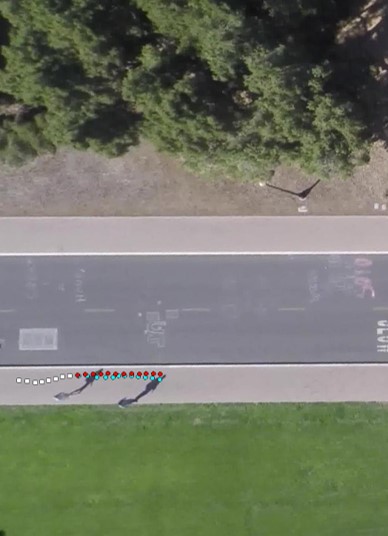}}
    \hfill
    \subfigure{\includegraphics[width=0.24\linewidth]{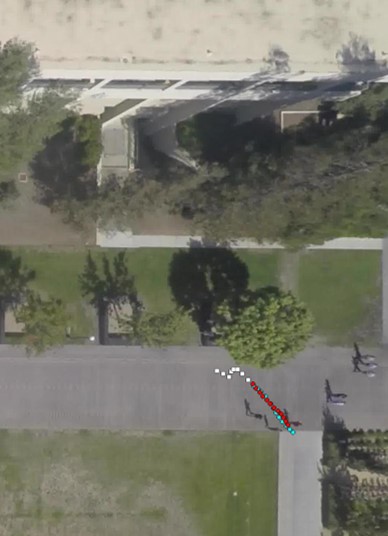}} \\
    \subfigure{\includegraphics[width=0.24\linewidth]{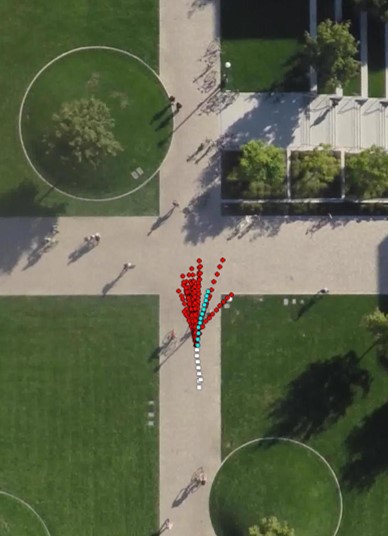}}
    \hfill
    \subfigure{\includegraphics[width=0.24\linewidth]{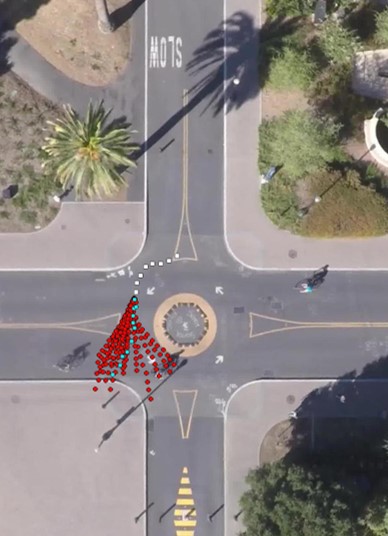}}
    \hfill
    \subfigure{\includegraphics[width=0.24\linewidth]{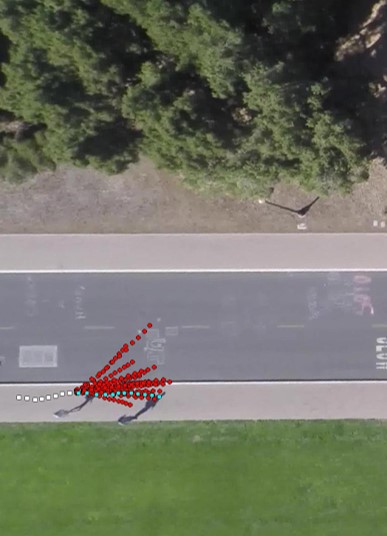}}
    \hfill
    \subfigure{\includegraphics[width=0.24\linewidth]{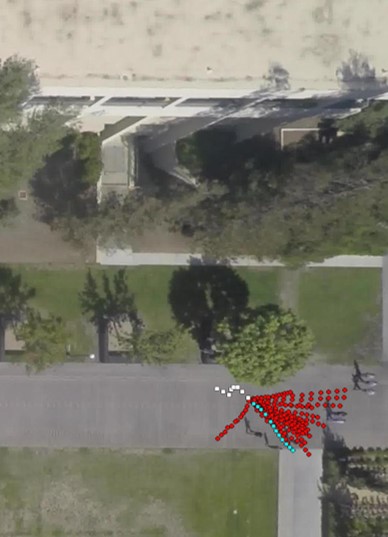}} \\
    \subfigure{\includegraphics[width=0.24\linewidth]{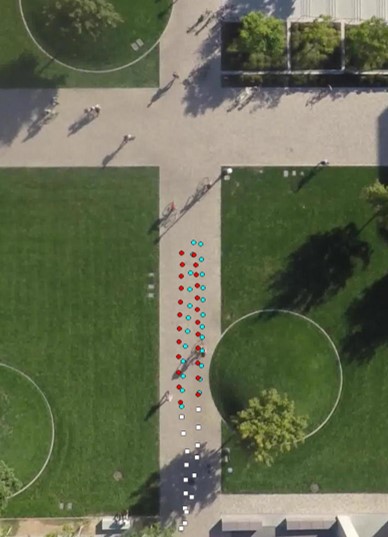}}
    \hfill
    \subfigure{\includegraphics[width=0.24\linewidth]{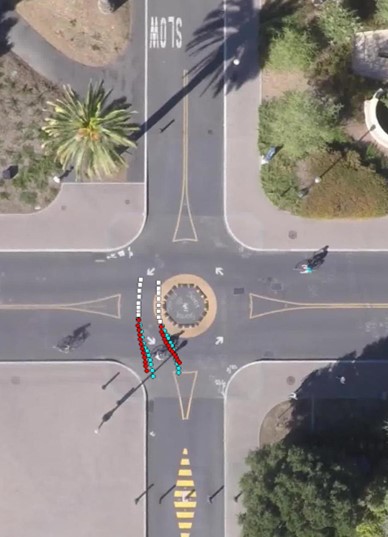}}
    \hfill
    \subfigure{\includegraphics[width=0.24\linewidth]{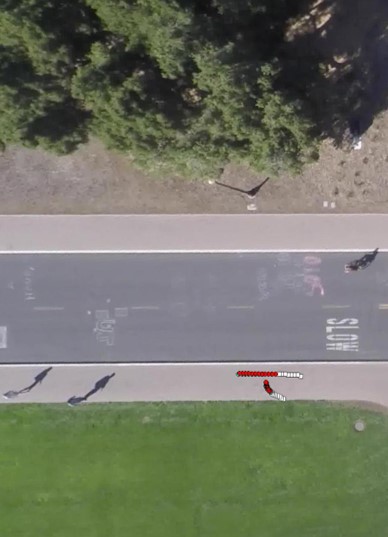}}
    \hfill
    \subfigure{\includegraphics[width=0.24\linewidth]{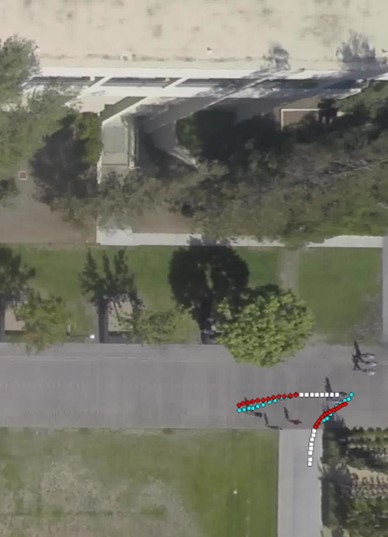}}
    \caption{Qualitative results of our proposed method across 4 different scenarios in the Stanford Drone. First row: The best prediction result sampled from 20 trials from LB-EBM. Second row: The 20 predicted trajectories sampled from LB-EBM. Third row: prediction results of agent pairs that has social interactions. The observed trajectories, ground truth predictions and our model's predictions are displayed in terms of white, blue and red dots respectively.}
    \label{fig:qual_results}
\end{figure*}

\textbf{Stanford Drone Dataset}: Table \ref{tab:sdd} summarizes the results of our proposed method against the baselines and state-of-the-art methods. Our proposed method achieves a superior performance compared to the previous state-of-the-art models \cite{bhattacharyya2019conditional,deo2020trajectory, mangalam2020not} on ADE by a significant margin of $10.9 \%$. While our improvement over other baselines on FDE is clear, the improvement over the PECNet is not significant. This might be because the PECNet focuses on optimizing the goal or the final step.   
\\

\begin{table*}[!htbp]
\centering
\begin{tabular}{c||c|c|c|c|c|c}
 &  ETH & HOTEL & UNIV & ZARA1 & ZARA2 & AVG\\\hline
\midrule
Linear *~\cite{alahi2016social}  & 1.33 / 2.94 & 0.39 / 0.72 & 0.82 / 1.59 & 0.62 / 1.21 & 0.77 / 1.48 & 0.79 / 1.59\\\hline
SR-LSTM-2 * ~\cite{zhang2019sr}  & 0.63 / 1.25 & 0.37 / 0.74 & 0.51 / 1.10 & 0.41 / 0.90 & 0.32 / 0.70 & 0.45 / 0.94\\\hline
S-LSTM~\cite{alahi2016social}  & 1.09 / 2.35 & 0.79 / 1.76 & 0.67 / 1.40 & 0.47 / 1.00 & 0.56 / 1.17 & 0.72 / 1.54\\\hline
S-GAN-P~\cite{gupta2018social}  & 0.87 / 1.62 & 0.67 / 1.37 & 0.76 / 1.52 & 0.35 / 0.68 & 0.42 / 0.84 & 0.61 / 1.21\\\hline
SoPhie~\cite{sadeghian2019sophie}  & 0.70 / 1.43 & 0.76 / 1.67 & 0.54 / 1.24 & 0.30 / 0.63 & 0.38 / 0.78 & 0.54 / 1.15\\\hline
MATF~\cite{zhao2019multi}  & 0.81 / 1.52 & 0.67 / 1.37 & 0.60 / 1.26 & 0.34 / 0.68 & 0.42 / 0.84 & 0.57 / 1.13\\\hline
CGNS~\cite{li2019conditional}  & 0.62 / 1.40 & 0.70 / 0.93 & 0.48 / 1.22 & 0.32 / 0.59 & 0.35 / 0.71 & 0.49 / 0.97\\\hline
PIF~\cite{liang2019peeking} & 0.73 / 1.65 & 0.30 / 0.59 & 0.60 / 1.27 & 0.38 / 0.81 & 0.31 / 0.68 & 0.46 / 1.00\\\hline
STSGN~\cite{zhang2019stochastic}  & 0.75 / 1.63 & 0.63 / 1.01 & 0.48 / 1.08 & 0.30 / 0.65 & 0.26 / 0.57 & 0.48 / 0.99\\\hline
GAT~\cite{kosaraju2019social} & 0.68 / 1.29 & 0.68 / 1.40 & 0.57 / 1.29 & 0.29 / 0.60 & 0.37 / 0.75 & 0.52 / 1.07\\\hline
Social-BiGAT~\cite{kosaraju2019social} & 0.69 / 1.29 & 0.49 / 1.01 & 0.55 / 1.32 & 0.30 / 0.62 & 0.36 / 0.75 & 0.48 / 1.00\\\hline
Social-STGCNN~\cite{mohamed2020social} &0.64 / 1.11 &0.49 / 0.85 &0.44 / 0.79 & 0.34 / 0.53 &0.30 / 0.48 & 0.44 / 0.75\\\hline
PECNet~\cite{mangalam2020not} &0.54 / 0.87 &0.18 / 0.24 &0.35 / 0.60 & 0.22 / 0.39 &0.17 / 0.30 &0.29 / 0.48\\\hline

\midrule
\textbf{Ours} &\textbf{0.30} / \textbf{0.52} & \textbf{0.13} / \textbf{0.20} &\textbf{0.27} / \textbf{0.52} & \textbf{0.20} / \textbf{0.37} &\textbf{0.15} / \textbf{0.29} &\textbf{ \textbf{0.21}} / \textbf{0.38}\\\hline
\end{tabular}
\caption{ADE / FDE metrics on ETH-UCY for the proposed LB-EBM and baselines are shown. The models with * mark are non-probabilistic. All models use 8 frames as history and predict the next 12 frames. Our model achieves the best average error on both ADE and FDE metrics. The lower the better.}
\label{tab:eth-ucy}
\end{table*}

\textbf{ETH-UCY}: Table \ref{tab:eth-ucy} shows the results for the evaluation of our proposed method on the ETH/UCY scenes. We use the leave-one-out evaluation protocol following CGNS \cite{li2019conditional} and Social-GAN \cite{gupta2018social}. We observe that the proposed LB-EBM outperforms prior methods, including the previous state-of-the-art \cite{li2019conditional}. We improve over the state-of-the-art on the average ADE by $27.6\%$ with the effect being the most on ETH ($44.4\%$) and least on ZARA1 ($9.1\%$). We also observe a clear improvement on the FDE.

\subsection{Qualitative Results}
In this section, we present qualitative results of our proposed method on the Stanford Drone dataset. In Figure~\ref{fig:qual_results}, we inspect the results under three different setups across 4 different scenarios. Those scenarios are selected involving various road conditions including crossing, sidewalk and roundabout. The first row presents the best prediction result, among 20 random samples drawn from the LB-EBM with respect to the ADE criterion, for each scenario. Our model is able to produce predictions that are close to the ground-truth trajectories in these scenarios. The second row illustrates the 20 predicted trajectories sampled from our method. By visualizing the results, we can see that LB-EBM is able to generate multi-modal and diverse predictions. Further, we display the prediction results of a pair of agents with social interactions in the third row. Interaction details such as ``straight going together'', ``turning together'', ``yielding'' and ``collision avoidance'' are captured by our proposed model. It demonstrates the effectiveness of our LB-EBM to model the agent-wise interactions for trajectory predictions.

\subsection{Ablation Study}
We conduct ablation studies to examine the important components of our model. In particular, we ablate each component of the overall learning objective as specified in Equation~\ref{eq:recon} - \ref{eq:kl2}. The results are summarized in Table~\ref{tab:obj_ablation}. Equation~\ref{eq:recon} is the basic reconstruction term and has to be kept. But we can replace Equation~\ref{eq:recon} and \ref{eq:pred} with ${\rm E}_{q_\phi(\pmb{Z}|\pmb{Y}, \pmb{X})} \log p(\pmb{Y} |\pmb{Z}, \pmb{X})$. That is, the model predicts the full trajectory directly without generating a plan first. It is corresponding to \textit{EBM without Plan} in Table~\ref{tab:obj_ablation}. Equation~\ref{eq:kl1} and \ref{eq:kl2} together are the KL divergence between the variational posterior $q_\phi(\pmb{Z}|\pmb{P}, \pmb{X})$ and the EBM prior $p_\alpha(\pmb{Z} | \pmb{X})$  (note that $p_0(\pmb{Z})$ is the base distribution for the EBM). We can replace $p_\alpha(\pmb{Z} | \pmb{X})$ with a Gaussian distribution conditional on $\pmb{X}$, corresponding to the \textit{Gaussian with Plan} condition. The previous two changes together lead to the \textit{Gaussian without Plan} condition. The ablation results indicate the effectiveness of the latent belief EBM and two-step approach.

In addition, we evaluate the model without the social pooling such that LB-EBM makes predictions only based on an agent's own action history (see the \textit{EBM with Plan without Social} condition in Table~\ref{tab:obj_ablation}). The decreased performance in ADE and FDE of this condition indicates that LB-EBM is effective to take into account social cues when provided.  

\begin{table}[!htbp]
\centering
\begin{tabular}{c||c|c}
 Time Steps &  ADE & FDE \\\hline
\midrule
Gaussian without Plan  & 18.61 & 27.55 \\\hline
EBM without Plan  & 10.28 & 18.60 \\\hline
Gaussian with Plan  & 9.53 & 16.32 \\\hline
EBM with Plan without Social  & 9.23 & 16.57 \\\hline
EBM with Plan  & 8.87 & 15.61 \\\hline

\end{tabular}
\caption{ADE / FDE metrics on Stanford Drone for different ablation conditions. The lower the better.}
\label{tab:obj_ablation}
\end{table}

\section{Conclusion}
In this work, we present the LB-EBM for diverse human trajectory forecast. LB-EBM is a probabilistic cost function in the latent space accounting for movement history and social context. The low-dimensionality of the latent space and the high expressivity of the EBM make it easy for the model to capture the multimodality of pedestrian trajectory distributions.  LB-EBM is learned from expert demonstrations (i.e., human trajectories) projected into the latent space. Sampling from or optimizing the learned LB-EBM is able to yield a social-aware belief vector which is used to make a path plan. It then helps to predict a long-range trajectory. The effectiveness of LB-EBM and the two-step approach are supported by strong empirical results. Our model is able to make accurate, multimodal, and social compliant trajectory predictions and improves over prior state-of-the-arts performance on the Stanford Drone trajectory prediction benchmark by $10.9\%$ and on the ETH-UCY benchmark by $27.6\%$.

\section*{Acknowledgment}
 
The work is supported by NSF DMS-2015577 and DARPA XAI project N66001-17-2-4029. 

\newpage
{\small
\bibliographystyle{ieee_fullname}
\bibliography{main.bib}
}

\clearpage

\section*{Appendix A: Learning}

\subsection*{Model Formulation}

Recall that $\pmb{X} = \{\pmb{x}_i, i = 1,...,n\}$ indicates the past trajectories of all agents in the scene. Similarly, $\pmb{Y}$ indicates all future trajectories. $\pmb{Z}$ represents the latent belief of agents. $\pmb{P}$ denotes the plans. We model the following generative model,
\begin{equation}
p_{\psi}(\pmb{Z}, \pmb{P}, \pmb{Y} | \pmb{X}) = \underbrace{p_{\alpha}(\pmb{Z} | \pmb{X})}_\text{LB-EBM} \overbrace{p_{\beta}(\pmb{P} |\pmb{Z}, \pmb{X})}^\text{Plan} \underbrace{p_{\gamma}(\pmb{Y} |\pmb{P},\pmb{X})}_\text{Prediction}. 
\end{equation}

\subsection*{Maximum Likelihood Learning}
Let $q_{data}(\bp, \by | \bx) q_{data}(\bx)$ be the data distribution that generates the (multi-agent) trajectory example, $(\bp, \by, \bx)$, in a single scene. The learning of parameters $\psi$ of the generative model $p_{\psi}(\bz, \bp, \by | \bx)$ can be based on $\text{min}_\psi \kl{q_{data}(\bp, \by|\bx)}{p_{\psi}(\bp, \by | \bx)}$ where $\kl{q(x)}{p(x)} = \E_q[\log q(x)/p(x)]$ is the Kullback-Leibler divergence between $q$ and $p$ (or from $q$ to $p$ since $\kl{q(x)}{p(x)}$ is asymmetric). If we observe training examples $\{(\bp_j, \by_j, \bx_j), j=1,..,N\} \sim q_{data}(\bp, \by | \bx) q_{data}(\bx)$, the above minimization can be approximated by maximizing the log-likelihood, 
\begin{align}
\sum_{j=1}^N \log p_{\psi}(\pmb{P}_j, \pmb{Y}_j | \pmb{X}_j) &= \sum_{j=1}^N \log \int_{\pmb{Z}_j} p_{\psi} (\pmb{Z}_j, \pmb{P}_j, \pmb{Y}_j | \pmb{X}_j)\label{eq:mle} 
\end{align}
which leads to the maximum likelihood estimate (MLE). Then the gradient of the log-likelihood of a single scene can be computed according to the following identity,
\begin{align}
&\nabla_{\psi} \log p_{\psi}(\pmb{P}, \pmb{Y} | \pmb{X}) = \frac{1}{p_{\psi}(\pmb{P}, \pmb{Y} | \pmb{X})} \nabla_{\psi} \int_{\pmb{Z}} p_{\psi} (\pmb{Z}, \pmb{P}, \pmb{Y} | \pmb{X})  \\
&= \int_{\pmb{Z}} \frac{p_{\psi} (\pmb{Z}, \pmb{P}, \pmb{Y} | \pmb{X})}{p_{\psi} (\pmb{P}, \pmb{Y} | \pmb{X})} \nabla_{\psi} \log p_{\psi} (\pmb{Z}, \pmb{P}, \pmb{Y} | \pmb{X}) \\
&= \int_{\pmb{Z}} \frac{p_{\psi} (\pmb{Z}| \pmb{X})p_{\psi} (\pmb{P} |\pmb{Z}, \pmb{X}) p_{\psi} (\pmb{Y} | \pmb{P},\pmb{X})}{p_{\psi} (\pmb{P} | \pmb{X}) p_{\psi} (\pmb{Y} | \pmb{P}, \pmb{X})} \nabla_{\psi} \log p_{\psi} (\pmb{Z}, \pmb{P}, \pmb{Y} | \pmb{X}) \\
& = \E_{p_{\psi}(\pmb{Z} | \pmb{P}, \pmb{X})} \nabla_{\psi} \log p_{\psi} (\pmb{Z}, \pmb{P}, \pmb{Y} | \pmb{X}).
\end{align}
The above expectation involves the posterior $p_{\psi}(\pmb{Z} | \pmb{P}, \pmb{X})$ which is however intractable. 

\subsection*{Variational Learning}
Due to the intractiablity of the maximum likelihood learning, we derive a tractable variational objective. Define 
\begin{equation}
q_{\phi}(\bz, \bp, \by | \bx) = q_{data}(\bp, \by|\bx)q_{\phi}(\bz | \bp, \bx)
\end{equation}
where $q_{\phi}(\bz | \bp, \bx)$ is a tractable variational distribution, particularly, a Gaussian with a diagnoal covariance matrix used in this work. Then our variational objective is defined to be the tractable KL divergence below, 
\begin{equation}
\kl{q_{\phi}(\bz, \bp, \by | \bx)}{p_{\psi}(\bz, \bp, \by | \bx)}
\end{equation}
where $q_{\phi}(\bz, \bp, \by | \bx)$ involves either the data distribution or the tractable variational distribution. Notice that,    
\begin{align}
&\kl{q_{\phi}(\bz, \bp, \by | \bx)}{p_{\psi}(\bz, \bp, \by | \bx)} \\
&= \kl{q_{data}(\bp, \by|\bx)}{p_{\psi}(\bp, \by | \bx)} \\
&+ \kl{q_{\phi}(\bz | \bp, \bx)}{p_{\psi}(\bz |\bp, \bx)} \\
\end{align}
which is an upper bound of $\kl{q_{data}(\bp, \by|\bx)}{p_{\psi}(\bp, \by | \bx)}$ due to the non-negativity of KL divergence, in particular, $\kl{q_{\phi}(\bz | \bp, \bx)}{p_{\psi}(\bz |\bp, \bx)}$, and equivalently a lower bound of the log-likelihood.

We next unpack the generative model $p_{\psi}(\bz, \bp, \by | \bx)$ and have,

\begingroup\makeatletter\def\f@size{7}\check@mathfonts
\def\maketag@@@#1{\hbox{\m@th\large\normalfont#1}}%
\begin{align}
&\kl{q_{\phi}(\bz, \bp, \by | \bx)}{p_{\psi}(\bz, \bp, \by | \bx)} \\
&= \kl{q_{data}(\bp, \by | \bx) q_{\phi}(\bz | \bp, \bx)}{p_{\alpha}(\bz | \bx) p_{\beta}(\bp | \bz, \bx) p_{\gamma}(\by | \bp, \bx)}\\
&= \E_{q_{data}(\bx)} \E_{q_{data}(\bp, \by|\bx) q_{\phi}(\bz | \bp, \bx)} \log \frac{q_{\phi}(\bz | \bp, \bx)} {p_{\alpha}(\bz | \bx)} \label{eq:ebm}\\
&+ \E_{q_{data}(\bx)}\E_{q_{data}(\bp, \by|\bx) q_{\phi}(\bz | \bp, \bx)} \log \frac{q_{data}(\bp | \by, \bx)}  {p_{\beta}(\bp | \bz, \bx)} \label{eq:plan}\\
&+ \E_{q_{data}(\bx)}\E_{q_{data}(\bp, \by|\bx) q_{\phi}(\bz | \bp, \bx)} \log \frac {q_{data}(\by | \bx )}  {p_{\gamma}(\by | \bp, \bx)} \label{eq:prediction}
\end{align}\endgroup 
Expressions \ref{eq:ebm}, \ref{eq:plan}, \ref{eq:prediction} are the major objectives for learning the LB-EBM, plan, and prediction modules respectively. They are the "major" but not "only" ones since the whole network is trained end-to-end and gradients from one module can flow to the other. We next unpack each of the objectives (where $\E_{q_{data}(\bx)}$ is omitted for notational simplicity).
 
Expression \ref{eq:ebm} drives the learning of the LB-EBM.
\begingroup\makeatletter\def\f@size{7}\check@mathfonts
\def\maketag@@@#1{\hbox{\m@th\large\normalfont#1}}%
\begin{align}
&\E_{q_{data}(\bp, \by|\bx) q_{\phi}(\bz | \bp, \bx)} \log \frac{q_{\phi}(\bz | \bp, \bx)} {p_{\alpha}(\bz | \bx)}\\
&= \E_{q_{data}(\bp, \by|\bx) q_{\phi}(\bz | \bp, \bx)} \log \frac{q_{\phi}(\bz | \bp, \bx)} {p_0 (\bz) \exp[- C_{\alpha}(\bz, \bx)]/Z_{\alpha}(\bx) } \\
& = \kl{q_{\phi}(\bz | \bp, \bx)}{p_0 (\bz)} \label{infer_reg}\\
& + \E_{q_{data}(\bp, \by|\bx) q_{\phi}(\bz | \bp, \bx)} C_{\alpha}(\bz, \bx) + \log Z_{\alpha}(\bx) \label{eq:ebm_learning}
\end{align}\endgroup 
where ${\scriptstyle Z_\alpha(\bx) = \int_{\bz} \exp(-C_\alpha (\bz, \bx)) p_0(\bz) = \E_{p_0(\bz)}(-C_\alpha (\bz, \bx))}$. \\

Let $\mathcal{J}(\alpha) = \E_{q_{data}(\bx)}\E_{q_{data}(\bp, \by|\bx) q_{\phi}(\bz | \bp, \bx)} C_{\alpha}(\bz, \bx) + \E_{q_{data}(\bx)} \log Z_{\alpha}(\bx)$, which is the objective for LB-EBM learning and follows the philosophy of IRL. And its gradient is,
\begin{align}
&\nabla_\alpha \mathcal{J}(\alpha) \\
&= \E_{q_{data}(\bx)}\E_{q_{data}(\bp, \by | \bx) q_\phi (\bz | \bp, \bx)} [\nabla_\alpha C_\alpha(\bz, \bx)] \label{eq:positive}\\
&- \E_{q_{data}(\bx)}\E_{p_\alpha (\bz | \bx)}[\nabla_\alpha C_\alpha (\bz, \bx)] \label{eq:negative}
\end{align}
Thus, $\alpha$ is learned based on the distributional difference between the expert beliefs and those sampled from the current LB-EBM. The expectations over $q_{data}(\bx)$ and $q_{data}(\bp, \by | \bx)$ are approximated with a mini-batch from the empirical data distribution. The expectation over $q_\phi (\bz | \bp, \bx)$ is approximated with samples from the variational distribution through the reparameterization trick. The expectation over $p_\alpha (\bz | \bx)$ is approximated with samples from Langevin dynamics guided by the current cost function.   

Expression \ref{eq:plan} drives the learning of the plan module. 
\begin{align}
(\ref{eq:plan}) &= -\E_{q_{data}(\bx)}\E_{q_{data}(\bp, \by|\bx) q_{\phi}(\bz | \bp, \bx)} \log p_{\beta}(\bp | \bz, \bx)\\ 
&- H(\bp | \by, \bx)
\end{align}
where $H(\bp | \by, \bx)$ is the conditional entropy of $q_{data}(\bp|\bx, \by)$ and is a constant with respect to the model parameters. Thus minimizing \ref{eq:plan} is equivalent to maximizing the log-likelihood of $p_{\beta}(\bp | \bz, \bx)$. 

Expression \ref{eq:prediction} drives the learning of the prediction module. 
\begin{align}
(\ref{eq:prediction}) &= -\E_{q_{data}(\bx)}\E_{q_{data}(\bp, \by|\bx) q_{\phi}(\bz | \bp, \bx)} \log p_{\gamma}(\by | \bp, \bx) \label{eq:pred_mle}\\ 
&- H(\by | \bx)
\end{align}
where $H(\by | \bx)$ is the conditional entropy of $q_{data}(\by | \bx)$ and is constant with respect to the model parameters. We can minimize Expression \ref{eq:pred_mle} for optimizing the prediction module. In the learning, $\bp$ is sampled from the data distribution $q_{data}(\bp, \by|\bx)$. In practice, we find sampling $\bp$ from the generative model $p_{\beta}(\bp | \bz, \bx)$ instead facilitates learning of other modules, leading to improved performance. The objective for learning the prediction module then becomes,
\begin{equation}
-\E_{q_{data}(\bx)}\E_{q_{data}(\by | \bx)}\E_{q_{\phi}(\bz | \bx)} \E_{p_{\beta}(\bp |\bz, \bx)} \log p_{\gamma}(\by | \bp, \bx)
\end{equation}
where 
\begin{align}
&\E_{q_{\phi}(\bz | \bx)} \\
&= \int_{\bp} q_{data}(\bp | \by, \bx) q_\phi (\bz | \bp, \bx) \\
&= \E_{q_{data}(\bp | \by, \bx)} q_\phi (\bz | \bp, \bx).
\end{align}

\section*{Appendix B: Negative Log-Likelihood Evaluation}
Although Best-of-K on ADE and FDE (e.g., $K=20$) is widely-adopted \cite{gupta2018social, kosaraju2019social, mangalam2020not, zhao2019multi}, some researchers \cite{ivanovic2019trajectron, salzmann2020trajectron++, thiede2019analyzing} recently propose to use kernel density estimate-based negative log likelihood (KDE NLL) to evaluate trajectory prediction models. This metric computes the negative log-likelihood of the groud-truth trajectory at each time step with kernel density estimates and then averages over all time steps. We compare the proposed LB-EBM to previous works with published results on NLL. They are displayed in Table~\ref{tab:nll}. Our model performs better than S-GAN \cite{gupta2018social} and Trajectron \cite{ivanovic2019trajectron} but underperforms Trajectron++\footnote{Trajectron++ is a concurrent work to ours and was discovered in the reviewing process.} \cite{salzmann2020trajectron++}. It might be because Trajectron++ use a bivariate Gaussian mixture to model the output distribution, while our model employs a unimomal Gaussian following most previous works. Our model can also be extended to adopt Gaussian mixture as the output distribution and we leave it for future work. 

\begin{table}[!htbp]
\centering
\begin{tabular}{c||c|c|c|c|}
 &  S-GAN & Trajectron & Trajectron++ & Ours\\\hline
\midrule
ETH  & 15.70  & 2.99 & 1.80 & 2.34 \\\hline
Hotel & 8.10 & 2.26 & -1.29 & -1.16 \\\hline
Univ  & 2.88  & 1.05 & -0.89 & 0.54 \\\hline
Zara1 & 1.36 & 1.86 & -1.13 & -0.17 \\\hline
Zara2 & 0.96 & 0.81 & -2.19 & -1.58 \\\hline
Average & 5.80  & 1.79 & -0.74 & -0.01 \\\hline
\end{tabular}
\caption{NLL Evaluation on ETH-UCY for the proposed LB-EBM and baselines are shown. The lower the better.}
\label{tab:nll}
\end{table}

\end{document}